\documentclass[letterpaper, 10 pt, conference]{ieeeconf}  % Comment this line out if you need a4paper

\IEEEoverridecommandlockouts                              % This command is only needed if 
\usepackage{graphics} %

\usepackage{epsfig} %
\usepackage{mathptmx} %
\usepackage{times} %
\usepackage{amsmath} %
\usepackage{amssymb}  %
\usepackage{multirow}
\usepackage{color}
\usepackage{amsfonts}
\usepackage{xcolor}
\usepackage{tikz}
\usepackage{etex}
\usepackage{url}
\usepackage{footnote}
\usepackage{floatrow}
\usepackage{booktabs}
\usepackage{inputenc}
\usepackage{txfonts}
\usepackage{multicol}
\usepackage{siunitx}
\usepackage{rotating}
\usepackage{arydshln}
\usepackage{algorithm}
\usepackage{algorithmic}
\usepackage{subcaption}
\usepackage[noadjust]{cite} % Added by ET for Ref issue

\title{\LARGE \bf
Towards Physically Talented Aerial Robots with Tactically Smart Swarm Behavior thereof: An Efficient Co-design Approach
}

\author{Prajit KrisshnaKumar$^{1}$, Steve Paul$^{1}$, Hemanth Manjunatha$^{1}$,  Mary Corra$^{2}$, \\ Ehsan Esfahani$^{1}$ and Souma Chowdhury$^{1,2*}$% <-this % stops a space
\thanks{$^\dagger$ Corresponding Author, soumacho@buffalo.edu}
\thanks{$^{1}$ Mechanical \& Aerospace Eng. Dep.,
        University at Buffalo, Buffalo, NY 
        }%
\thanks{$^{2}$ Computer Science and Eng. Dep.,
University at Buffalo, Buffalo, NY }%
\thanks{*This work was supported by the NSF award CMMI 2048020 and the ONR grant N00014-24-1-2003. Any opinions, findings, conclusions, or recommendations expressed in this paper are those of the authors and do not necessarily reflect the views of NSF and/or ONR.}% Any opinions, findings, conclusions, or recommendations expressed in this paper are those of the authors and do not necessarily reflect the views of NSF}
\thanks{Copyright \textcopyright 2024 ASME. Personal use of this material is permitted. Permission from ASME must be obtained for all other uses, in any current or future media, including reprinting/republishing this material for advertising or promotional purposes, creating new collective works, for resale or redistribution to servers or lists, or reuse of any copyrighted component of this work in other works}
}

\begin{document}

\maketitle
\thispagestyle{empty}
\pagestyle{empty}

%%%%%%%%%%%%%%%%%%%%%%%%%%%%%%%%%%%%%%%%%%%%%%%%%%%%%%%%%%%%%%%%%%%%%%%%%%%%%%%%
\begin{abstract}
The collective performance or capacity of collaborative autonomous systems such as a swarm of robots is jointly influenced by the morphology and the behavior of individual systems in that collective. In that context, this paper explores how morphology impacts the learned tactical behavior of unmanned aerial/ground robots performing reconnaissance and search \& rescue. This is achieved by presenting a computationally efficient framework to solve this otherwise challenging problem of jointly optimizing the morphology and tactical behavior of swarm robots. Key novel developments to this end include the use of physical talent metrics and modification of graph reinforcement learning architectures to allow joint learning of the swarm tactical policy and the talent metrics (search speed, flight range, and cruising speed) that constrain mobility and object/victim search capabilities of the aerial robots executing these tactics. Implementation of this co-design approach is supported by advancements to an open-source Pybullet-based swarm simulator that allows the use of variable aerial asset capabilities. The results of the co-design are observed to outperform those of tactics learning with a fixed Pareto design, when compared in terms of mission performance metrics. Significant differences in morphology and learned behavior are also observed by comparing the baseline design and the co-design outcomes.

\end{abstract}

%%%%%%%%%%%%%%%%%%%%%%%%%%%%%%%%%%%%%%%%%%%%%%%%%%%%%%%%%%%%%%%%%%%%%%%%%%%%%%%%
\section{Introduction}
Collective intelligence enables a swarm of robotic systems to adapt effectively to uncertain and unknown environments, autonomously organize themselves, and exhibit emergent behaviors that lead to superior problem-solving capabilities.
Through the collaborative efforts of multiple robots working in tandem, tasks (e.g., exploration, transportation, surveying, harvesting, search \& rescue, and assembly of distributed objects~\cite{cheraghi2022past}) that are beyond the capabilities of any single robot can be efficiently tackled. 
% In nature, collective intelligence is displayed by organisms such as bees, ants, and birds. These organisms accomplish complex tasks like intricate nest construction without central oversight, relying on decentralized, simple behavioral rules followed by each individual members of the group. This gave rise to a new field of swarm robotics, where a team of robots act in decentralized manner to achieve a common goal, that is challenging for an individual robot. 
% Swarm robotics holds significant potential across a wide array of real-world applications~\cite{cheraghi2022past}, particularly in tasks such as exploration, transportation, surveying, harvesting, search \& rescue, and assembly of distributed objects. 
Yet, realizing the potential of swarm robotics and collective intelligence involves addressing formidable challenges with respect to design choices that shape the operating envelope and functionalities of the individual members of a swarm or multi-robot team.
For example, consider the generic swarm scenario in which individual robots in a swarm autonomously assess their surroundings, communicate findings with each other, and collaboratively plan and execute future tasks or actions. 
The challenge is that there are no clear-cut, principled approaches to designing the low-level behaviors (individual decisions or policies) and individual robot morphology that will ensure the desired collective behaviors.

Emergent behavior in swarm robotics/collective intelligence results from simple rules followed by each entity and their interaction with each other and their environment \cite{chen2022securing}. These interactions give rise to complex and adaptive behaviors that are robust and efficient. 
However, this emergent behavior cannot be directly inferred from an individual's behavior or capabilities; rather, it is a product of dynamic interplay within the swarm. Minor modifications in the design of individual robots might affect the robot's operating envelope, significantly impacting its emergent behavior at the collective level. 
Most often, \textit{behavior} is trained or developed based on fixed or apriori-designed physical systems \cite{ hachiya2022reinforcement, wang2017autonomous}.
This approach inherently restricts each robot's operational capabilities, often resulting in designs that do not fully optimize performance and thus limit the overall effectiveness of the swarm.
The intricate interplay between \textit{morphology} (physical form/design) and \textit{behavior} (e.g., robot's decisions that enable coordinated motion and task completion) must be optimized together to explore how efficiently the swarm as a whole can perform desired operations without failure. 
This co-design process ensures that physical design and behavioral algorithms evolve together, facilitating the alignment of capabilities to achieve superior collective performance. 
%In this paper, we propose a computational framework that enables co-optimization of the morphology and behavior of individual robots in a swarm with machine learning-based policies/behavior to achieve maximum performance from its emergent behavior.

Machine learning-based policies are becoming increasingly popular in expressing perception and control/planning loops in robots and autonomous systems, including those that can work as a collaborative group. There has been some work on co-design for individual robots on control side~\cite{sims1994evolving,lipson2000automatic,weel2014robotic,khazanov2013exploiting,cheney2013unshackling,komosinski2009evolving,bongard2011morphological, zardini2022task, bergonti2023co, bravo2020one}. In the area of co-design with ML-based policies, Gupta et al., \cite{gupta2021embodied} introduced the DERL framework to create embodied agents for a complex animal morphology, which uses both traditional evolutionary methods and RL methods in parallel. 
Many of the earlier methods in this field relied on evolutionary algorithms, which can suffer from computational inefficiency. Among others, notable methods that are closest to our work are introduced by Schaff et al. \cite{Schaff2018JointlyLT}, who introduced a Deep Reinforcement Learning method for co-designing agents' morphology and behavior. In their method, an additional distribution for designs is introduced, and its parameters are updated to maximize the policy reward. Another method that is closest to our work is introduced by Luck et al. \cite{luck2020data}, where four individual networks, two for morphology and two for behavior, are trained in parallel. These works are showcased in singular robotics environments using PyBullet, focusing on control systems with continuous state-action spaces. Firstly, most of these existing approaches seek to directly operate on the raw morphological (design) space, which becomes computationally prohibitive as the complexity of the system increases. What is thus currently lacking is the understanding of how morphology affects (usually a smaller set of) fundamental or latent system capabilities, which in turn constrain or shape the envelope of feasible behaviors and exploit this understanding to decompose the co-design problem into a sequence of simpler search problems. Secondly, unlike in classical co-design, there exists very little work on systematic co-design of swarm or multi-robot systems.
%Despite these advancements in individual robotics systems, there remains a significant gap in research concerning co-design in swarm robotics systems, particularly in exploring the optimal morphology and behavior choices that drive the collective emergent behavior. 

%However, in real-world autonomous systems, the complexity of the system can significantly expand the morphology (variable) space, and this expansion results in prohibitive computational demands. 

To address these gaps, in this paper, we propose a computational framework that enables the concurrent design of i) the morphology of individual robots in a swarm and ii) their collective behavior. Here, we utilize our previously proposed artificial-life-inspired talent metrics \cite{zeng2022efficient} that are physical quantities of interest, reflective of the capabilities of an individual robotic system (e.g., range, nominal power consumption, weight, sensing FoV, payload capacity, turning radius, etc.). 
Talent metrics represent a compact yet physically interpretable parametric space that connects the behavior space and morphology space. We use this to decompose the morphology-behavior co-optimization into a sequence of talent-behavior optimization problems that can effectively reduce the overall search space (for each individual problem) without negligible compromise in the ability to find optimal solutions. In other words, the decomposition approach presented here is nearly lossless, i.e., a solution that can be found otherwise with a brute-force nested optimization approach to co-design will also exist in the overall search space spanned by our decomposed co-design approach (albeit assuming that each search process is ideal). We also propose a novel talent-infused actor-critic method to optimize the talents and learn the behavior concurrently. 
%Further, we consider multi-robot systems composed of complex unmanned aerial vehicles (UAVs) designed to execute real-world tasks. While the application considered here is multi-robot based, the proposed framework is general across any Reinforcement Learning (RL) behavior for embodied intelligence. 

To demonstrate the efficacy of the proposed co-design approach, we examine its application in a complex Urban Search and Rescue operation using heterogeneous swarm robots. We call this problem, \textbf{SWA}rm robotic search and \textbf{R}escue \textbf{M}ission for \textbf{C}omplex \textbf{A}dversarial \textbf{E}nvironment (SWARM-CAE). Often, in complex robotics missions, it is imperative to combine multiple swarm behaviors to accomplish a higher-level goal. Behjat et al., \cite{behjat2021learning} introduced a tactical learning framework for complex swarm missions where the higher-level goals are decomposed into multiple sub-goals that are achieved by combining individual swarm and single-robot behaviors; further, this framework also provides abstraction methods to overcome the state and action space explosion in multi-robot systems pertinent to learning based methods. Due to the framework's versatile nature, it is adapted for our SWARM-CAE problem.  Here, we consider an urban or semi-urban environment where the operation takes place, with the robots spawned at a nearby depot location. The robots are commanded by a neural network-based policy that decides the tactical behavior (aka allocation of different tasks and GoTo locations) to rescue victims or find/extract objects of interest from the environment as quickly as possible while mitigating the loss of swarm agents to adversarial entities in the environment -- these characteristics are typical of disaster response, humanitarian missions, and planetary explorations. The neural network-based policy guiding the behavior of the swarm is trained using reinforcement learning (RL) over experience collected in an open-source robot simulator. 
This application was chosen because it effectively showcases the benefits of co-designing swarm systems, particularly in managing the complexities and dynamics of real-world missions where multiple swarm behaviors are required to achieve higher-level goals. Often, the swarm search and rescue frameworks available currently are restricted to grid-based environments and lack real-world characteristics. The application presented here takes place in real-world maps obtained with OpenStreetMaps API and simulated with SHaSTA (an Open Source Simulator) \cite{manjunatha2024shasta}.

%The expected collective behavior should exhibit a tactical strategy to rescue the victim as soon as possible while avoiding adversities. %In this study, we use a reinforcement learning (RL)-based methodology that concurrently learns tactical behaviors and tailors robot talents, all while respecting morphological constraints. We compare our results of co-designed behavior with behavior learned using a fixed design chosen from among the best trade-offs. 

\textit{Thus, the primary contribution of this paper is a computational framework capable of concurrently learning the talents (driven by morphology) and behavior of RL-guided swarm robotic systems performing real-world missions.} The secondary objective is to formulate a Markov Decision Process (MDP) on top of the graph for the SWARM-CAE Problem and Graph Capsule Convolution network (GCAPCN) to serve as the tactical policy network for the MDP. 
Though the assumed application uses a heterogeneous team of two different types of robots, unmanned aerial vehicles (UAVs) and unmanned ground vehicles (UGVs), in this paper, for simplicity, we only optimize the morphology of UAVs. In this proposed approach, first, we derive talent metrics for quadcopter-type UAVs using a set of logical principles based on the application, perform multi-objective optimization to obtain the Pareto front in talent space, and then create a regression surface representation of this talent Pareto. The talent Pareto is used in our proposed novel Talent-infused Actor-Critic approach to optimize for mission efficiency. Finally, we perform another optimization to find the exact morphology corresponding to the learned talents. 

The remaining portion of this paper is organized as follows: in section \ref{sec2}, we define the co-design problem using an example, thereby defining the concept of talent metrics; concurrently, we provide an overview of our proposed talent-infused actor-critic method in a generalized form; 
section \ref{sec3} introduces the SWARM-CAE problem; section \ref{sec:GRL} describes the graph neural network developed for the proposed Talent-infused Actor-Critic method; section \ref{sec5} presents the results of using Talent-infused actor-critic method on the SWARM-CAE Problem, compared to a baseline; finally in section \ref{sec:conclusion}, we present our conclusions.  

%%%%%%%%%%%%%%%%%%%%%%%%%%%%%%%%%%%%%%%%%%%%%%%%%%%%%%%%%%%%%%%%%%%%%%%%%%%

\section{Framework for Concurrent Design of Behavior and Morphology}
\label{sec2}\
Consider a group of UAVs performing a search operation over an environment. Let $\mathbf{X}_M$ represent the vector of morphological variables of individual UAVs, such as wing span, frame, or battery capacity. This vector encompasses all the variables necessary to build a complete system. 
Meanwhile, $\boldsymbol\Phi$ represents controllable parameters of the algorithm that guide its behavior. Depending on the complexity, $\boldsymbol\Phi$ can be a single heuristic parameter or the neural network's weights if the behavior is based on it. 
The performance metric, denoted as $f_L$, quantifies its effectiveness in the environment. In the context of RL, this can be a reward function. 
The primary goal of co-design optimization in this context is to maximize this performance metric, and this can be represented as an optimization problem shown in Eq.~\eqref{eq:objfun_morph_1}. 

\begin{equation}
\begin{aligned}
&\textrm{Max:} & & f_L(\mathbf{X}_M, \boldsymbol\Phi )\\
&\textrm{S. t.:} & &\mathbf{X}_{\texttt{min}} \leq \mathbf{X}_M \leq \mathbf{X}_{\texttt{max}} \\
& & &\boldsymbol\Phi_{\texttt{min}} \leq \boldsymbol\Phi \leq \boldsymbol\Phi_{\texttt{max}} \\
& & & g(\mathbf{X}_{{M}}) \leq 0 
\end{aligned}
\label{eq:objfun_morph_1}
\end{equation}

\begin{figure*}
    \centering
    \includegraphics[width=0.9\linewidth]{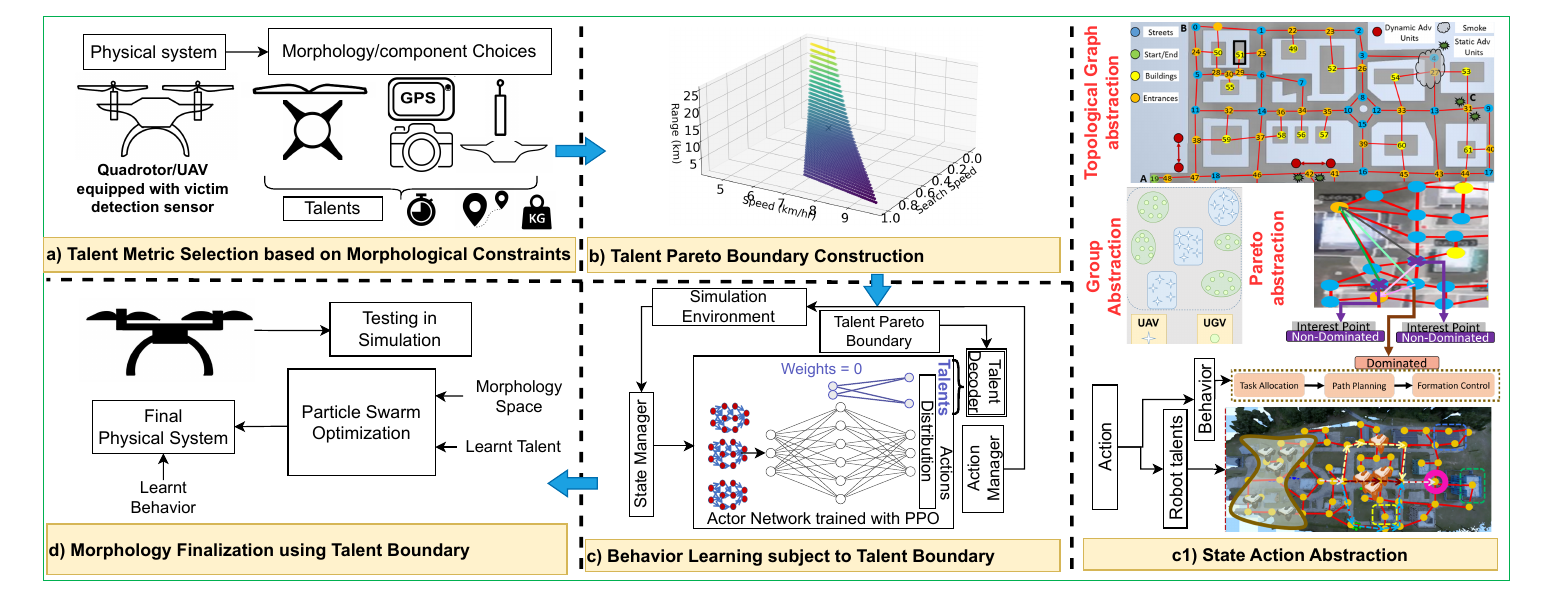}
    \caption{Flowchart of our co-design framework, a) Morphology and its dependent talent parameters are derived, b) Based on the talents, we create a Pareto boundary, c) The Talent-infused Actor-critic method is used to train the associated behavior and talents, d) Finalize the morphology for the optimized talents.}
    \label{fig:overal_frame_work_new}
\end{figure*}

Three primary methods are used to solve this optimization problem: Sequential Design: first optimize the Morphology $\mathbf{X}_M$ and optimize for the behavior $\boldsymbol\Phi$ or vice versa; this leads to a highly sub-optimal design/collective behavior and cannot be generalized \cite{yu2023multi}; Nested Design: optimize both the behavior and morphology in a nested way, while it can be thorough, it is computationally intensive \cite{christensen2010anatomy}; Evolutionary methods: use evolutionary optimization methods to optimize the behavior and morphology together, this approach is computationally feasible when the behavior or morphology is simple, as the complexity increases, the computational cost increases \cite{chee2014simultaneous}. 
Using the talent metrics concept proposed in our previous paper \cite{zeng2022efficient}, we decompose the morphology-behavior design into a series of sequential talent-based optimization problems. 
Figure \ref{fig:overal_frame_work_new} shows the four steps involved in our proposed co-design framework. The below subsections delve deep into each step. 

\subsection{Morphology Constrained Talent Metrics Selection}
The talent metrics ($\mathbf{Y}_{\texttt{TL}}$) refer to the morphology-driven variables that directly influence the resultant behavior's performance. 
For instance, given different combinations of morphology variables, we obtain different metrics representing the UAV's operating envelope, such as flight range and cruising speed. 
It is evident that these talent metrics form a parametric layer given by
\begin{equation}
    \mathbf{Y}_{\texttt{TL}}=f_M(\mathbf{X}_M)
\end{equation}  
where $f_M$ denotes the model responsible for determining the talents corresponding to a morphology. This can be obtained with simulations, supervised learning if a dataset is available, and analytical equations. $\mathbf{Y}_{\texttt{TL}}$ is a vector consisting of values [$\text{Y}_{\texttt{TL,1}}, \ldots,\text{Y}_{\texttt{TL,m}}$]. Upon introducing talents, the objective function Eq.~\eqref{eq:objfun_morph_1} can be modified as an optimization problem expressed in Eq.~\eqref{eq:objfun_morph_2}.

\begin{equation}
\begin{aligned}
&\textrm{Max:} & & f_L(\mathbf{Y}_{TL}, \boldsymbol\Phi)\\
&\textrm{S. t.:} & &\mathbf{Y}_{TL}, \boldsymbol\Phi \in \mathbf{R} \\
& & &\mathbf{Y}_{TL_{\texttt{min}}} \leq \mathbf{Y}_{TL} \leq \mathbf{Y}_{TL_{\texttt{max}}} \\
& & &\boldsymbol\Phi_{\texttt{min}} \leq \boldsymbol\Phi \leq \boldsymbol\Phi_{\texttt{max}} \\
\end{aligned}
\label{eq:objfun_morph_2}
\end{equation}

To effectively replace morphology with talents, the selected talent metrics should satisfy four principles: 
1) The collection of talent metrics should depend only on morphology. 
2) Talent metrics should exhibit the monotonic goodness property, meaning that for each metric, there should be a direction (increasing or decreasing) corresponding to improved performance. 
3) Talent metrics should be collectively exhaustive in determining the impact on the performance of the behavior, meaning there cannot be a case where constraints or bounds of behavior can change with a fixed talent. 
4) Each talent metric should conflict with the others; for example, increasing the payload reduces the range. 

Advantages of substituting $\mathbf{X}_{\text{M}}$ with $\mathbf{Y}_{\texttt{TL}}$ are i) Directly optimizing the talent space allows focusing on the most relevant performance metrics, this often leads to finding the optimal talents and behavior more efficiently than when co-optimizing morphology and behavior,  ii) provides a likely dimension reduction; Typically, the morphology space is considerably larger than the talent space; iii) the monotonous goodness property allows for safely eliminating solutions in the dominated region and constraining the optimization to Pareto front.

\subsection{Talent Pareto Boundary Construction}\label{subsec:talentpareto}
Given the monotonic goodness property of each talent metric, the optimal talent should lie within the Pareto region. Therefore, we perform a multi-objective optimization to identify the non-dominated (Pareto) points of the talent variables. The optimization problem can be expressed as
\begin{equation}
\begin{aligned}
&\textrm{Max:} & & (\text{Y}_{\texttt{TL,1}}, \ldots,\text{Y}_{\texttt{TL,m}}) = f_{M}(\mathbf{X}_{M})\\
&\textrm{S. t.:} & &\mathbf{X}_{\texttt{min}} \leq \mathbf{X}_M \leq \mathbf{X}_{\texttt{max}} \\
& & & g(\mathbf{X}_{M}) \leq 0 
\end{aligned}
\label{eq:objfun_morph_3}
\end{equation}
This exposes the Pareto points for the multi-objective optimization problem. 
The pareto front can be modeled through an approximation or surrogate method.
The model can be represented as
\begin{align}
    \text{Y}_{\texttt{TL},m} = f_{\text{S}} \left(\text{Y}_{\texttt{TL},1},\ldots,\text{Y}_{\texttt{TL},m-1} \right)
    \label{eq:surrugate_model}
\end{align}
 where $f_{\text{S}}$ represents the approximation model, and $m$ represents the quantity of talents metrics.

\subsection{Behavior Learning via Talent-infused Actor-critic subject to Talent Boundary}
In the context of a group of UAVs performing the search example provided before, consider a complex neural network is used for the behavior of individual robots and trained through an actor-critic-based RL algorithm. The Actor-critic algorithm consists of 2 neural networks. The actor network maps the current state of the UAV to the actions, and it dictates how the UAV should behave. The actor policy can be denoted as $ \pi((a|s; \theta) $ where $\theta$ are the weights of the actor network and $a$ represents the action given state $s$. While the actor decides the actions given a state, the critic network evaluates the potential value (state-value) of being in that state, estimating the future rewards that can be obtained from that state. The value network can be represented as  $V(s; w)$ where $V$ is the value of state ($s$) with weights of the network ($w$). 
In order to solve our optimization problem shown in Eq.~\eqref{eq:objfun_morph_2}, we need to incorporate talents into the actor and critic network. The below sections delve deep into the modifications performed as well as the pseudo-code of the proposed Talent-infused Actor-critic Algorithm.

\subsubsection{Modifications in Actor-Critic Framework}
The actor network that typically generates the action/behavior for the UAV is augmented with an additional neural network called the talent network. Figure \ref{fig:overal_frame_work_new}c) provides an example representing this augmentation. 
The output shape of the talent network is $m-1$, where $m$ represents the total number of talents.  The weights of the input neurons of the talent network are set to zero, and not-trainable, i.e., the network doesn't require a state or input to work and always provides a constant output. 
These outputs directly correspond to the mean of $m-1$ distributions, which is further processed through the talent decoder to obtain the final talent values. Here, in order to limit the optimization inside the Pareto front, the final layer of the talent network is given a sigmoid activation function and is processed through a talent decoder. 
The policy of this actor network is given by $ \pi((a|s, \hat{Y}_{\texttt{TL,1}}, \cdots,\hat{Y}_{\texttt{TL,m-1}}); \theta) $, where $a$ is the behavioral action $\theta$ indicates the weights and $\hat{Y}_{\texttt{TL,1}}, \cdots,\hat{Y}_{\texttt{TL,m-1}}$ are the outputs from talent network.

\subsubsection{Talent Decoder}
The talent decoder's objective is to scale the values from the talent network to ensure they remain within the bounds of the talent Pareto front. 
Once the actor network provides the talent values  ($\hat{Y}_{\texttt{TL,1}}, \hat{Y}_{\texttt{TL, 2}}, .., \hat{Y}_{\texttt{TL, m-1}}$), we need to scale these values using upper and lower bounds to get the talent value ($\mathbf{Y}_{TL}$). In order to find the upper and lower limits of each talent value, we employ quantile Regression model at 5th and 95th percentile respectively conditioned on previous talents. For the First Talent ($\text{Y}_{TL,1}$), we can directly get the lower and upper limits. Hence, we use the below equation to obtain the first talent. 
\begin{align}
    \text{Y}_{TL,1} = \hat{Y}_{TL,1}(max(\hat{Y}_{TL,1}) - min(\hat{Y}_{TL,1}))  + min(\hat{Y}_{TL,1})
\end{align}
From the 2nd to $m-1$ talents, we use the following equation, 

\begin{align*}
    \text{Y}_{\text{TL},i} = &\ \hat{Y}_{\text{TL},i} \left( Q(0.95|\text{Y}_{\text{TL},1}, \ldots, \text{Y}_{\text{TL},i-1}) \right. \\
    &\ \left. - Q(0.05|\text{Y}_{\text{TL},1}, \ldots, \text{Y}_{\text{TL},i-1}) \right) + \\
    &\ Q(0.05|\text{Y}_{\text{TL},1}, \ldots, \text{Y}_{\text{TL},i-1}) \\
    &\ \forall i \in \{2, \ldots, m-1\}
\end{align*}

This scaling allows us to stay within the Pareto front. The scaled values are passed onto the simulation for creating the robots. 

\subsubsection{Training Phase}
Here, the main objective is to optimize the distribution with the talent network and learn the behavior. 
During the first step of each episode, We do a forward pass in the actor network, which is then followed by sampling through distribution. The augmented output of the actor network can be given by 
\begin{align}
    A_\theta(s_{i}) &= (a_{i}, \hat{Y}_{\text{TL,1}}, \hat{Y}_{\text{TL, 2}}, \ldots, \hat{Y}_{\text{TL, m-1}}), 
    \quad \text{for all } i \in \{1, \ldots, T\}
    \label{eq:actor_policy}
\end{align}
where $A_\theta(s_{(i)})$ signifies the output of actor policy at timestep $i$ with input state $s_{(i)}$. $a_{(i)}$ represents the action for state $s_{(i)}$, $\hat{Y}_{\texttt{TL,1}}, \hat{Y}_{\texttt{TL, 2}}, .., \hat{Y}_{\texttt{TL, m-1}}$ represents the talent values from 1 to $m-1$. 
These $m-1$ values are subsequently processed by a talent decoder, which scales these values based on the maximum and minimum bounds of their respective talents. To get the final talent $\hat{Y}_{\texttt{TL, m}}$, we use the approximation model created with Eq.~\eqref{eq:surrugate_model}.
Following the determination of actions and talents, these values are fed into the simulation environment. Based on these talents, robots are instantiated, and the chosen action is executed, which then returns us with the reward and the new states. The new states are passed on to the actor network again. Crucially, after the first step of the episode, talent values are not sampled from the talent network and the actor network continues to suggest actions based on the current state without any changes until the episode ends. Essentially, talents stay the same throughout an episode, but the states and actions update with each step, as shown in the Eq.~\eqref{eq:actor_policy}.

The critic network updates primarily using the state value, incorporating an additional component called "talents" into the state space. This integration results in the input to the network containing both state and talent values. Instead of calculating a state value using the critic network, the critic network now calculates the state-talent value pair. Consider $V(s_t,\hat{Y}_{\text{TL}}; w)$ represents the value of state from critic network with weights $w$ for states $s_t$ and all the talents from actor network $\hat{Y}_{\text{TL}}$
The talents employed by the UAV during the episode are used in the state-talent input for the critic network. 
The critic assesses the state value based on these talents, offering an estimate of the expected future reward accumulation for the given talent and actions with state $s$. The Temporal Difference (TD) error is computed based on 
\begin{align}
    \delta = r + \gamma V(s_{(t+1)},\hat{Y}_{\text{TL}} ; w) - V(s_t, \hat{Y}_{\text{TL}}; w) 
\end{align}
The critic network updates its weight $w$ to minimize the TD error and the update rule can be represented as 
\begin{align}
    w_{t+1} = w_t + \alpha \delta \nabla_w V(s_t, \hat{Y}_{\text{TL}}; w)
\end{align}
If the cumulative reward from the actor's actions and talents exceeds the critic's estimate, the critic provides positive feedback, encouraging the actor to increase the probability of the current action and associated talent.
\begin{align}
    \nabla_\theta J(\theta) = \delta \nabla_\theta \log \pi(a|s_t, \hat{Y}_{\text{TL}}; \theta)
\end{align}

Over time, this iterative process enables the actor to refine both its policy and talent parameters, striving for an optimal balance that maximizes cumulative rewards. To ensure generalizability, updates should be performed over a large batch of episodes. This necessity arises from accumulating gradients over a diverse set of talent values.

The pseudo-code for talent-infused actor-critic method is shown in Alg.\ref{alg:talent_infused_ac}. Most of the open-source RL libraries, including Stable-baselines3 \cite{stable-baselines3} and OpenAI Baselines \cite{baselines}, follow updates over the batch method. In libraries, parallel vectorized environments are created to collect experiences and update the policy after collecting a batch of episodes. For more details on the implementations, please refer to \cite{shengyi2022the37implementation}.

\subsubsection{Testing Phase}
During the testing phase, the distributions are removed, thereby taking deterministic actions. Since the talent network (part of the actor network) has weights of 0 in the input layer, the state space doesn't affect the outcome, and it always results in a single value. These final values passed through the talent decoder will provide us with the optimized talents $\mathbf{Y}_{\texttt{TL}}^*$, and the learned policy is the result of the optimization problem expressed in Eq.~\eqref{eq:objfun_morph_2}.  Let us consider the optimized talents as $\mathbf{Y}_{\texttt{TL}}^*$.

\begin{algorithm}[!ht]
\scriptsize
\caption{Talent-Infused Actor-Critic Method}
\label{alg:talent_infused_ac}
\begin{algorithmic}[1]
\STATE \textbf{Input:} Learning rates $\alpha_\theta$ and $\alpha_w$, discount factor $\gamma$, batch size $B$, and the total number of talent variables $m$
\STATE Initialize actor network $A_\theta$ with weights $\theta$, outputting policy $\pi((a, \mathbf{Y}_{\text{TL}})|s; \theta)$, where $\mathbf{Y}_{\text{TL}}$ are the talent values and $a$ is the behavioral action
\COMMENT{Morphology-dependent weights are set to 0 and are non-trainable}
\STATE Initialize critic network with weights $w$, estimating value function $V(s, \hat{Y}_{\text{TL}}; w)$
\STATE Initialize experience buffer $\mathcal{E}$
\WHILE{not reached end of training}
    \FOR{$b = 1$ \TO $B$}
        \STATE Initialize start state $s_{(t_1)}$ for the $b$-th episode
        \STATE Obtain $a_{(t_1)}$ and $\hat{Y}_{\text{TL,1}}, \hat{Y}_{\text{TL,2}}, \ldots, \hat{Y}_{\text{TL, m-1}}$ from $A_\theta(s_{(t_1)})$
        \STATE Calculate $\hat{Y}_{\text{TL,m}}$ using $f_{\text{SM}}(\hat{Y}_{\text{TL,1}},\ldots,\hat{Y}_{\text{TL,m-1}})$
        \WHILE{not done}
            \STATE Use $a_{t_i}$ and $\hat{Y}_{\text{TL}}$ for simulation, where $t_i$ denotes the current timestep
            \STATE Execute action $a_{t_i}$, observe reward $r$ and next state $s_{(t+1)}$
            \STATE Store transition $(s_t, a_{t_i}, r, s_{(t+1)}, \hat{Y}_{\text{TL}})$ in $\mathcal{E}$
            \STATE For $t_i > t_1$, retain $\hat{Y}_{\text{TL}}$ without re-sampling
            \STATE Update $s_t$ to $s_{(t+1)}$
        \ENDWHILE
    \ENDFOR
    \FORALL{transitions $(s_t, a, r, s_{(t+1)}, \hat{Y}_{\text{TL}})$ in $\mathcal{E}$}
        \STATE Compute TD error: $\delta = r + \gamma V(s_{(t+1)},\hat{Y}_{\text{TL}} ; w) - V(s_t, \hat{Y}_{\text{TL}}; w)$
        \STATE Update critic weights: $w = w + \alpha_w \delta \nabla_w V(s_t, \hat{Y}_{\text{TL}}; w)$ 
        \STATE Update actor weights $\theta$ based on gradient: $\nabla_\theta J(\theta) = \delta \nabla_\theta \log \pi(a|s_t, \hat{Y}_{\text{TL}}; \theta)$
    \ENDFOR
    \STATE Clear experience buffer $\mathcal{E}$ for next batch
\ENDWHILE
\end{algorithmic}
\end{algorithm}

\subsection{Morphology Finalization using Learnt Behavior and Talent Boundary}
To finalize the morphology for optimized talents, another optimization approach given by Eq.~\eqref{eq:objfun_final_gen} should be handled
\begin{equation}
\begin{aligned}
&\textrm{Min:} & & f_f=||\mathbf{Y}_{\texttt{TL}}(\mathbf{X}_M)-\mathbf{Y}_{\texttt{TL}}^*|| \\
&\textrm{Subject to:} & &\mathbf{X}_M \in \mathbf{R} \\
& & &\mathbf{X}_{\texttt{min}} \leq \mathbf{X}_M \leq \mathbf{X}_{\texttt{min}} \\
% & & & = 0 \ \text{(if present)}\\
\end{aligned}
\label{eq:objfun_final_gen}
\end{equation}
where $f_f$ refers to the objective function value, and $\mathbf{Y}_{\texttt{TL}}^*$ represents the optimal talent. This optimization aims to obtain morphology that provides talents as close as possible to the optimal talents obtained in learning.

\section{Multi-robot Search and Rescue}\label{sec3}
We consider a swarm search and rescue operation happening in an urban environment involving UAVs and UGVs. Here, the robots aim to locate a specific building with an object of interest (victim) inside and rescue it. 
Initially, all robots form a preset platoon in a depot for deployment. The urban environment contains multiple suspect buildings (target buildings) where the victim can be, and the goal is to find the true target building (goal location) where the victim is. 
UAVs can search the building's perimeter to determine whether it is the true target building (goal location) with the exact location of the victim. Meanwhile, UGVs have indoor search capabilities, allowing them to search within the building and execute the rescue operation. 
If the UAVs successfully identify the target building and its location, the UGVs can bypass comprehensive exploration, expediting the rescue upon entry.
The outdoor search progress is calculated as $\psi_{l, \texttt{out}} = V_\psi \times (n_f \times P)$, considering the number of floors ($n_f$), total perimeter ($P$), and individual UAV search speed ($V_\psi$ or Field of view/FOV). The Search speed is based on a sensor that can detect the presence of a victim, and the quality of the sensor depends on its weight; the higher the weight of the sensor, the higher the $V_\psi$ value, and it allows faster perimeter search. 

Three types of adversaries are considered: Smoke, Bombs, and Dynamic adversaries. Dynamic adversaries follow fixed paths and continuously monitor the area. UGVs can neutralize dynamic adversaries but can also be neutralized. Bombs are not neutralizable and can destroy UGVs on contact. Smoke slows down UAVs but is undetectable by dynamic adversaries. We use an MDP formulation to solve the SWARM-CAE problem, focusing on optimal tactics for effective mission completion (see Section \ref{subsec:mdp} for details). To manage the state-action space, we employ encoding techniques, including group abstraction (robot platooning with various commands) and Pareto encoding of nodes (identifying critical points through non-dominated sampling). Further details can be found in~\cite{behjat2021learning}. Figure \ref{fig:overal_frame_work_new} c1) illustrates these abstractions and consists of a sample 3D environment where the mission is happening. A short video describing the environment and the mission can be found in this link\footnote{https://buffalo.box.com/s/3tadqfqtgv7jw5kcez7vt5gfc54ny2wq}

\subsection{Simulation}
We used a simulator called SHaSTA(Simulator for Human and Swarm Team Applications)~\cite{manjunatha2024shasta} for this study. SHaSTA has multiple advantages over other simulators, some of which are: 1) automated importing of any real-world maps using OpenStreetMap API and running swarm simulations, 2) inbuilt swarm-primitives such as formation control and path planning. Three different primitives are used: Task allocation, Path Planning, and Formation control. For path planning, we consider 3 different routes: i) \textbf{Aggressive path:} Fastest path to the destination, ii) \textbf{Normal path:} The path cannot have deadly adversaries such as bombs or dynamic adversaries. iii) \textbf{Cautious path:} No adversaries present. The policy model provides the tactical decision of location to search and the path to take to reach that location. The entire map is abstracted as a topological graph. The location of interests, such as buildings, intersections, and building entrances, are considered nodes of graphs. The path planning is done using the networkx library using this topological map. 
 The region-based formation control method~\cite{cheah2009region} is used here to navigate the platoons to the desired location. 
In order to implement our Co-design approach, a simulator must be able to import custom robots. New modules have been implemented for importing custom robots without completely closing the simulation. This helps collect experiences at a much faster pace and aids in learning faster.

\subsection{MDP Formulation}\label{subsec:mdp}
\subsubsection{States}\label{subsubsec:states}
SHaSTA allows importing any real-world location as a graph structure. 
This already solves the problem of discretizing a continuous environment. Further, not all locations on the map are equally important. Through Pareto optimality-based filtering, We identify the critical locations through a non-dominated sorting process. The Pareto filtering process can be given by $k^* = arg\min_{k} \ f_i (k)  =  P(G_l) \times t (X_k \rightarrow G_l), ~~ l=1,2,\ldots N_l$, where $X_k$ represents the spatial location of the $k$-th graph node that can be allocated as a destination; $t (X_k \rightarrow G_i)$ is the time taken to reach a potential target $G_i$ from the point $X_k$, and $P(G_l)$ is the probability that it is the true target $G_l$. At the beginning of the mission, the probability of each target building is set to $1/N_l$, and once either the indoor or outdoor search is completed for any target location, the probability of all the other locations gets updated. 

\begin{table}
\centering
\scriptsize
\caption{{States of the tactics model for a swarm with variable UAV Squads and UGV Squads}}
\label{tab:MDP_table}
\begin{tabular}{|l|l|l|} \hline
Graph & Property & Shape \\ 
\hline
\multirow{4}{*}{State of the mission} & Remaining Time & 1, \\
 & Remaining UAV platoons & 1, \\
 & Remaining UGV platoons & 1, \\ \cdashline{2-3}
 & shape & (3,1) \\ \hline
\multirow{5}{*}{UAV states $\mathcal{G}_{UAV}$} & Location & 2, \\
 & range & 1, \\
 & type & 1, \\
 & Goal Location & 1, \\ \cdashline{2-3}
 & shape & ($N_{UAV}$,5) \\ \hline
\multirow{6}{*}{UGV states $\mathcal{G}_{UGV}$} & Location & 2, \\
 & range & 1, \\
 & health & 1, \\
 & type & 1, \\ 
 & Goal Location & 1, \\ \cdashline{2-3}
 & shape & ($N_{UGV}$,5) \\ \hline
\multirow{5}{*}{Building states $\mathcal{G}_{BLD}$} & location & 2, \\
 & probability & 1, \\
 & indoor search progress & 1, \\
 & outdoor search progress & 1, \\ \cdashline{2-3}
 & shape  & ($N_{BLD}$,5) \\ \hline
\multirow{4}{*}{Acting Platoon $\mathbf{Y}_{ACT}$ } & location & 2, \\
 & type & 1, \\
 & range & 1, \\ \cdashline{2-3}
 & shape & 4,1 \\ \hline
\multirow{3}{*}{Adversary States $\mathcal{G}_{ADV}$} & location & 2, \\
 & type & 1, \\ \cdashline{2-3}
 & shape & ($N_{ADV}$,3) \\ \hline
 \multirow{4}{*}{UAV Talents $\mathbf{Y}_{\texttt{TL}}$} & Range & 1, \\
 & Velocity & 1, \\
 & Search Speed & 1, \\ \cdashline{2-3}
 & shape & 3,1 \\ \hline
\end{tabular}
\end{table}
We formulated 4 individual graphs for the states of UAV, UGV, Pareto node, and adversaries. Note that these graphs are input space for the Reinforcement Learning policy and differ from the environment graph used for abstraction and simulation. The UAV states are represented by $\mathcal{G}_{\text{UAV}} = (V_{\text{UAV}}, E_{\text{UAV}}, \Omega_{\text{UAV}})$, the UGV states are represented by $\mathcal{G}_{\text{UGV}} = (V_{\text{UGV}}, E_{\text{UGV}}, \Omega_{\text{UGV}})$, Pareto node states $\mathcal{G}_{\text{BLD}} = (V_{\text{BLD}}, E_{\text{BLD}}, \Omega_{\text{BLD}})$ and the adversary nodes  by $\mathcal{G}_{\text{ADV}} = (V_{\text{ADV}}, E_{\text{ADV}}, \Omega_{\text{ADV}})$. In each graph, $V$ represents the nodes/vertices specific to the state, $ E $ represents the edges connecting these vertices, and $ \Omega $ is the corresponding weighted adjacency matrix. Each node in these graphs represents an entity - be it a UAV, UGV, Pareto node, or adversary and each edge connects a pair of these nodes. The total count for each node type is denoted by $N_{UAV}, N_{UGV}, N_{BLD}$, and $N_{ADV}$, which represents the total number of UAVs, UGVs, Pareto nodes, and adversaries respectively. 

The complete state formulation is provided in table \ref{tab:MDP_table}. Apart from these graphs, we also include linear vectors to include the state of the mission, talents, and acting platoon in the state space. For the "state of the mission," we consider the remaining time and the current functional platoon counts. The UAV talent space consists of the talent vectors ($\mathbf{Y}_{\texttt{TL}}$). More details on the selection of talent metrics are explained in sec.\ref{sec5}. At each time step, an idle platoon is selected for which the action is required, and the properties of this platoon are given in the acting platoon vector.

\subsubsection{Actions}

The Behavioural action ($a$) here is discrete, and policy should decide which Pareto node to visit and the path to be taken (aggressive path, normal path, and cautious path) for the selected idle platoon. If the platoon runs out of range or health, it is considered non-functional and will not be considered further in the mission.

\subsubsection{Reward}
The proposed reward function takes into account the mission status, time, and casualties. 
If the scenario is successful, meaning the robots have rescued the victim within the allowed time, we provide rewards as follows
\begin{align}
    R = \tau_{sc} + (\Lambda_{sc})   \label{eq:rewards}
\end{align}
where ($\tau_{sc}$) is the rescue time that is the duration taken to rescue the victims and is normalized by the maximum allowed mission time. The survival rate ($\Lambda_{sc}$) represents the ratio of the number of robots that survive the mission to the initial size of the swarm. If the operation is not successful, we provide a negative reward of $-1$

 \begin{figure*}
    \centering
    \includegraphics[width=0.8\textwidth]{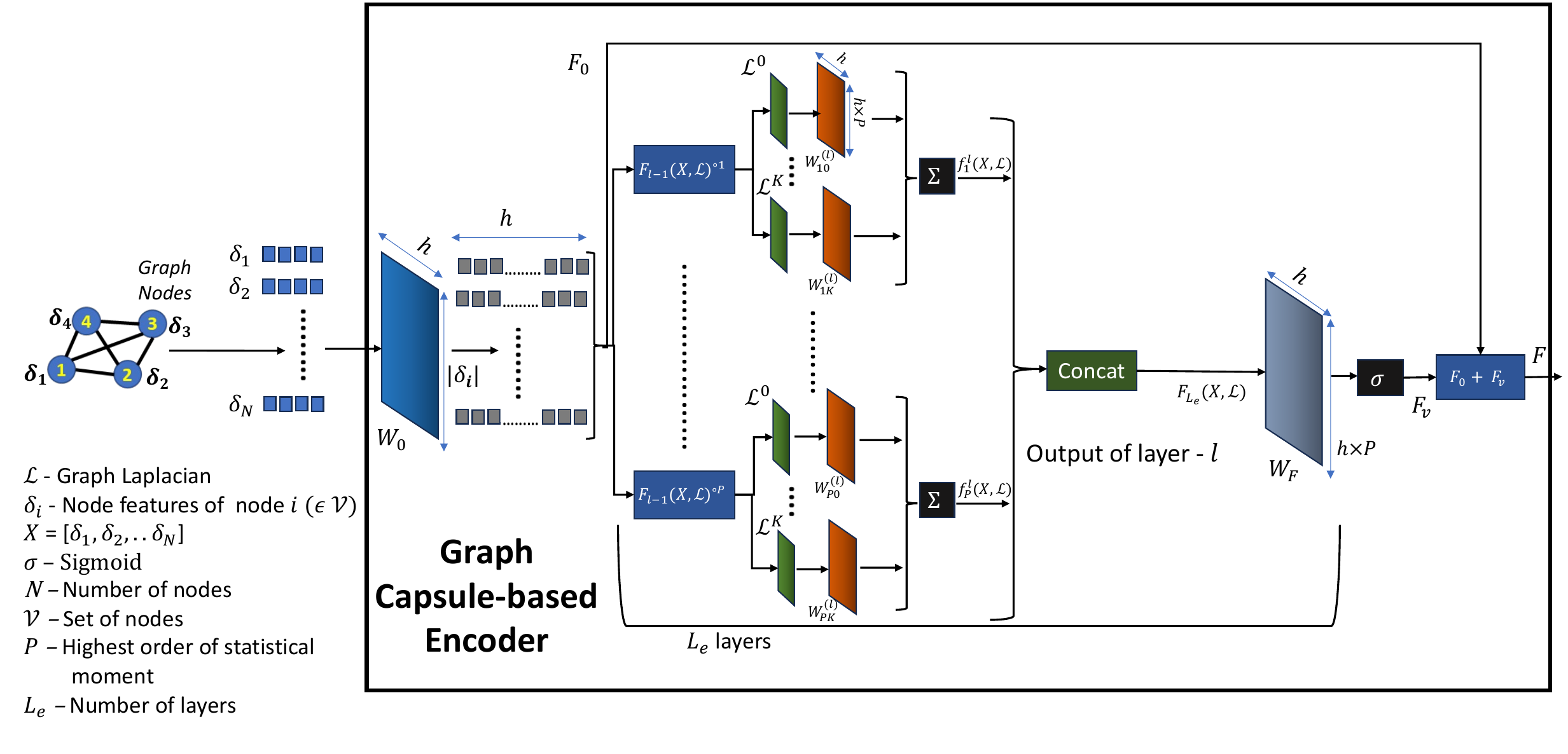}
    \caption{The GCAPCN-based encoder. The node properties undergo a linear transformation first, followed by multiple graph capsule layers.}
    \label{fig:gcaps}
\end{figure*}

\section{Graph based Reinforcement Learning}\label{sec:GRL}
The Reinforcement learning (RL) approach involves maximizing the total reward per episode by training a policy network to learn actions sequentially for the mission represented as an MDP whose objective is to maximize the total reward per episode. In this work, we implement a policy gradient-based on-policy method called Proximal Policy Optimization (PPO) \cite{schulman2017proximal}  to train the policy. During every decision-making instance, the policy takes in the state space variables and computes the action (in this case the which Pareto node to visit and the path to take, and also the talent metrics at the start of an episode.)
Since 4 of the main state space variables are represented as a graph (as explained in section \ref{subsec:mdp}), we develop a policy network based on Graph Neural Networks (GNN). The GNNs are used to compute node embeddings for the graph. In this work, we use Graph Capsule Convolutional Neural Networks (GCAPCN) \cite{Verma2018} as the GNN. GCAPCN has proved to be an excellent graph feature abstraction network (from our previous work on similar Multi-agent problems \cite{capam_mrta,paul2022scalable}) compared to other GNNs such as Graph Convolutional Networks (GCN), Graph Attention Networks (GAT), etc. We initialize 4 GCAPCN network for the 4 grpahs ($\mathcal{G}_{UAV}, \mathcal{G}_{UGV}, \mathcal{G}_{BLD}, \mathcal{G}_{ADV}$) respectively, and compute the corresponding node embeddings $F_{UAV} \in \mathbb{R}^{N_{UAV} \times h}$, $F_{UGV} \in \mathbb{R}^{N_{UGV}\times h}$, $F_{BLD} \in \mathbb{R}^{N_{BLD} \times h}$, and $F_{ADV} \in \mathbb{R}^{N_{ADV} \times h}$, respectively. Here $h$ is the embedding length. We compute feature vectors for representing the states of the acting platoon $F_{Act} \in \mathbb{R}^{1 \times h}$, and the UAV talents $F_{Tal} \in \mathbb{R}^{1 \times h}$, by two separate linear transformations (with learnable weights). The features $F_{UAV}$, $F_{UGV}$, $F_{ADV}$, $F_{ACT}$, and $F_{TAL}$ are used to compute a context vector $F_{context}$ (explained in section \ref{subsubsec:context_vec}). Since the goal of the policy is to select a Pareto node and a path (one out of three options), we compute logits for all the Pareto nodes across all three paths. We compute 3 logits vector ($LG_{P1} \in \mathbb{R}^{10 \times 1}$, $LG_{P2} \in \mathbb{R}^{10 \times 1}$, and $LG_{P3} \in \mathbb{R}^{10 \times 1}$), for the three types of path. We use 3 Multi-head Attention (MHA) based decoders to compute the logit vectors (explained in section \ref{subsubsec:mhadecoder}).

\subsection{Policy Model}
\subsubsection{Graph capsule-based feature abstraction}\label{subsubsec:GNN_encoder}
\begin{figure}[!ht]
    \centering
    \includegraphics[trim={0 2cm 0 0},clip,width=\linewidth]{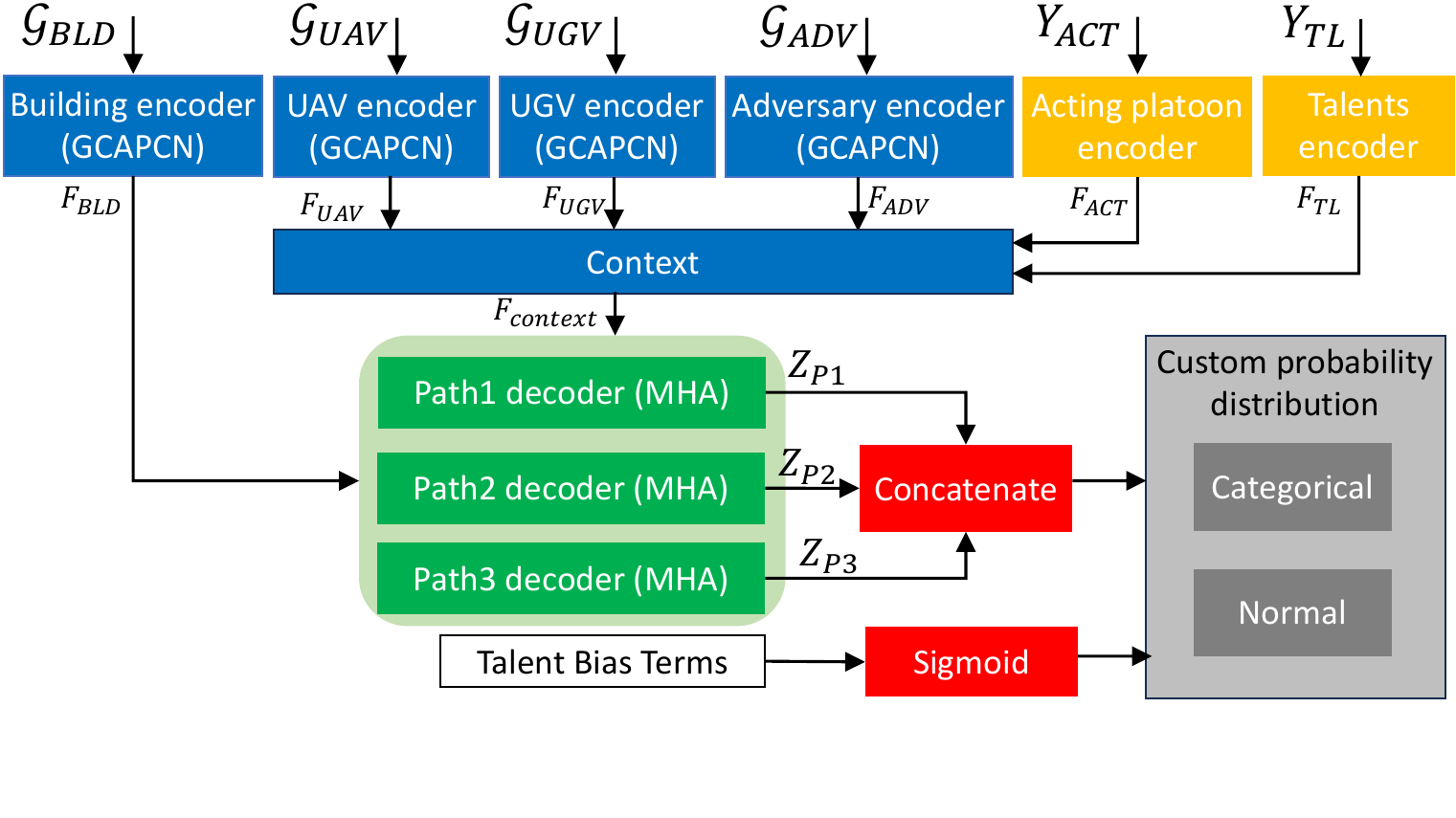}
    \footnotesize
    \caption{Structure of the overall policy model consisting of the GCAPCN encoders, Context, MHA-based decoders, Talent bias network, and the custom probability distribution.}
    \label{fig:policy_full}
\end{figure}

In order to compute a learned representation of the 4 graphs of the state space, we use a GCAPCN network to compute node embeddings. We initialize 4 GCAPCN networks for the 4 graphs. The GCAPCN networks (Fig. \ref{fig:gcaps}) take in a graph (in the form of node properties and the weighted adjacency matrix) and output the node embeddings. Here we give a very brief description of GCAPCN. Consider a graph $\mathcal{G} = (V, E, \Omega)$ with $N$ nodes, where $V$ is the set of nodes, $E$ is the set of edges, and $\Omega$ is the weighted adjacency matrix. Let $\delta_{i}$ represent the node properties of node $i \in V$, as a vector, and $X = [\delta_{1} \dots \delta_{N}] \in \mathbb{R}^{N \times |\delta_i|}$ be the node property matrix, where $|\delta_i|$ represents the cardinality of $\delta_i$. First, the node properties undergo a linear transformation $F_{0} \in \mathbb{R}^{N \times h}$, where $N$ is the number of nodes in the graph and $h$ is the embedding length. This is followed by multiple graph capsule layers \cite{Verma2018} that make use of the transformed node properties and the graph Laplacian matrix to compute permutation-invariant node embeddings $f^{p}_{l}(X,L) \in \mathbb{R}^{N \times h}, \ \forall \ p \in [1, P], l \in [1, L_{e}]$, where $P$ is the highest order of statistical moment and $L_{e}$ is the number of layers. This captures the nodal information (the node properties matrix $X$) and the structural information ( the Graph Laplacian $L$) of the nodes of the graph and has $P$ representations of this information. The node properties of the four graphs in the state space can be found in Table \ref{tab:MDP_table}. These embeddings are concatenated and done for multiple layers ($L_{e}$). The output from the final layer $F_{L_{e}(X,L)}$ is passed through a feedforward layer to get an embedding length of $h$, which is then added with $F_{0}$ to get the final embedding $F \in \mathbb{R}^{N \times h}$. For further information on GCAPCN, we refer the reader to \cite{Verma2018, capam_mrta}. In the main policy diagram (Fig. \ref{fig:policy_full}), the outputs are represented as $F_{BLD}$, $F_{UAV}$, $F_{UGV}$, and $F_{ADV}$
 \begin{figure}[h!]
    \centering
    \includegraphics[width=0.5\textwidth]{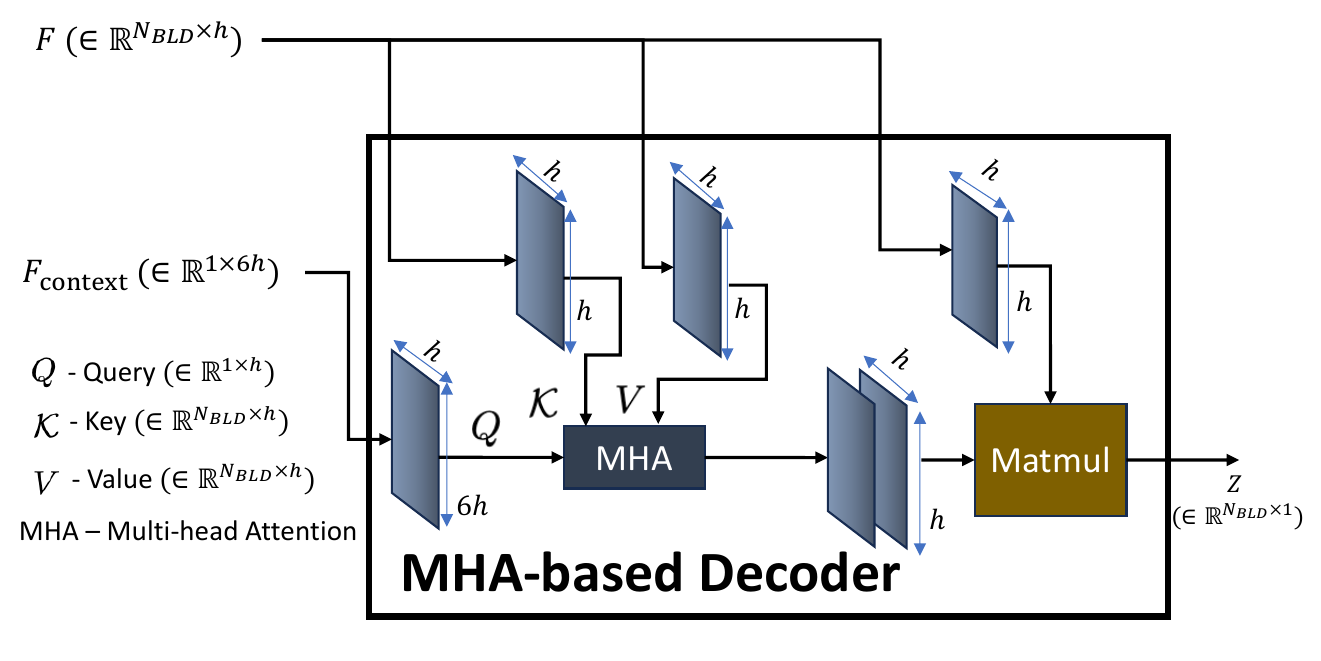}
    \footnotesize
    \caption{The MHA-based decoder. The input to the decoder includes the node embeddings and the context and the output is the computed logits.}
    \label{fig:mha_decoder}
\end{figure}
%  \begin{figure}[!ht]
%     \centering
%     \includegraphics[width=\textwidth]{figs/GCAPS.pdf}
%     \footnotesize
%     \caption{Structure of the GCAPCN encoder for computing the node embeddings of the graphs.}
%     \label{fig:gcaps}
% \end{figure}

%  \begin{figure}[!ht]
%     \centering
%     \includegraphics[width=\textwidth]{figs/MHA_decoder.pdf}
%     \footnotesize
%     \caption{Structure of the MHA-based decoder}
%     \label{fig:mha_decoder}
% \end{figure}
\subsubsection{Context Vector}\label{subsubsec:context_vec}
The context vector is computed using all the feature vectors. 
\begin{align}
    F_{context} = Concat(Mean(F_{BLD}), Mean(F_{UAV}), \\ \nonumber Mean(F_{UGV}), Mean(F_{ADV}), F_{ACT}, F_{TL}))
\end{align}
where $Mean(F_{BLD}) \in \mathbb{R}^{1 \times h}$ is the mean of the mean feature vector across all the nodes, and similarly for $Mean(F_{UAV})$, $Mean(F_{UGV})$, and $Mean(F_{ADV})$. The context vector $F_{context} \in \mathbb{R}^{1 \times 6h}$ will be used along with $F_{BLD}$ to compute the logits for the three paths using the MHA decoder (explained in the next section).

\subsubsection{Logits Computation using MHA-based Decoder}\label{subsubsec:mhadecoder}
As mentioned in the above section, given the current state, the goal is to select which Pareto node to visit and path to choose, and this selection is made based on computing 3 logits vectors ($Z_{P1}, Z_{P2}, Z_{P3}$). In order to compute the logits, we use an MHA-based decoder (Fig. \ref{fig:mha_decoder}). The decoder takes in the Pareto embeddings $F_{BLD}$ in the form of keys ($\mathcal{K}$) and values ($\mathcal{V}$), and the context $F_{context}$ in the form of query ($\mathcal{Q}$)
and computes compatibility scores between  $F_{context}$ and every node embedding in $F_{BLD}$ which is then used to compute the attention heads. These attention heads are then passed through a feedforward layer and multiplied with another linear transformation of the node embeddings to compute the final logits. This is done with three different encoders for computing logits for the three types of path, which is then used to compute the action distribution.

% \subsubsection{Action Distribution}
% Utilizing a custom distribution, values from the bias are passed to the sigmoid activation layer, which then we use Normal distribution from Pytorch to sample the talents. Here the policy produces 2 talent outputs, namely $\hat{Y}_{\texttt{TL,1}}$ and $\hat{Y}_{\texttt{TL,2}}$. The values of $\hat{Y}_{\texttt{TL,1}}$ is scaled the maximum and minimum possible values of  $\hat{Y}_{\texttt{TL,1}}$
% \begin{align}
%     \hat{Y}_{\texttt{TL,1}} &= \hat{Y}_{\texttt{TL,1}} \times (max_{\hat{Y}_{\texttt{TL,1}}} - min_{\hat{Y}_{\texttt{TL,1}}}) + min_{\hat{Y}_{\texttt{TL,1}}}
% \end{align}
% $\hat{Y}_{\texttt{TL,2}}$ is scaled based on the maximum and minimum values from the quantile regression model. $L_q, U_q$ is the lower and upper quantiles for quantile regression. 
% \begin{align}
%     L_{\hat{Y}_{\texttt{TL,2}}}, U_{\hat{Y}_{\texttt{TL,2}}} &= \text{QuantileRegression}(\hat{Y}_{\texttt{TL,1}}, L_q, U_q) \\
%         \hat{Y}_{\texttt{TL,2}} &= \hat{Y}_{\texttt{TL,2}} \times (U_{\hat{Y}_{\texttt{TL,2}}} - L_{\hat{Y}_{\texttt{TL,2}}}) + L_{\hat{Y}_{\texttt{TL,2}}};     \label{eqn:scaleytl2} 
% \end{align}

% These values are then input to the Pareto surrugate model to determine $\mathbf{Y}_{\texttt{TL,3}}$. 
% Furthermore, the probability values from the MHA-decoder layer are passed on to the Categorical Distribution to sample the behavior actions. 

\section{Case Study - Co-design for Swarm Tactics Learning}\label{sec5}

In this section, we showcase the results obtained by applying our proposed co-design framework to the SWARM-CAE problem. We delve deep into Talent selection based on the morphology of UAVs, Pareto model creation using polynomial regression, Co-learning using our proposed Talent-infused Actor-critic method, and finally, comparing the results of a co-design policy with a sequential design policy. 
\begin{figure}[h!]
    \centering
    \includegraphics[width=0.9\linewidth]{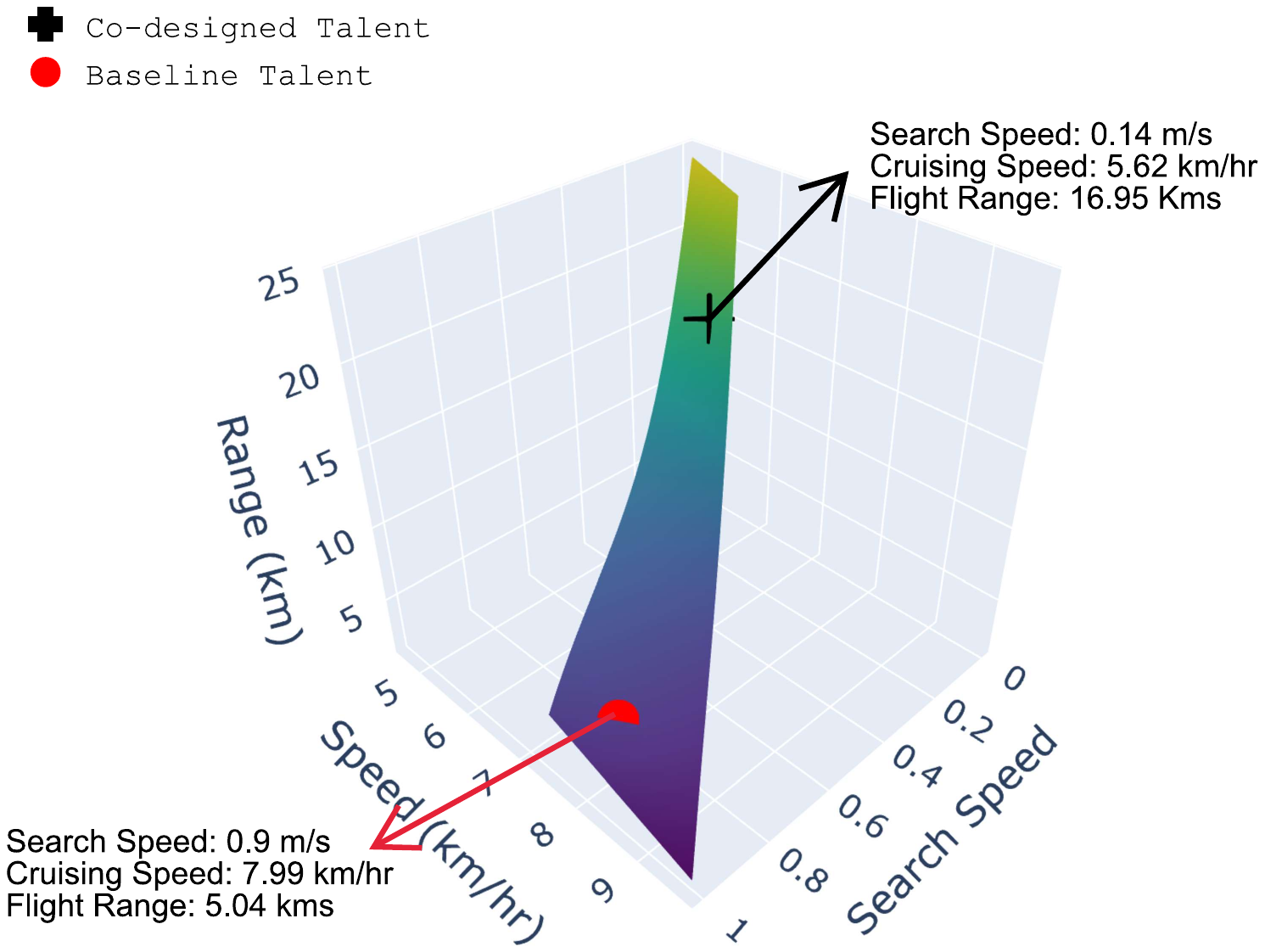}
    \caption{Talent Pareto front represented by Polynomial Regression applied to computed Pareto solutions obtained by multi-objective optimization of Talents; limits of talents captured with quantile regression.}
    \label{fig:paretomodel}
\end{figure}

\subsection{Talent Metrics and Pareto Model} \label{sec:sub_talents}
Here, we focus on developing a Blended-Wing-Body (BWB) integrated Quadcopter (BIQU) used in our previous studies \cite{zeng2022efficient}. Key morphological parameters that determine the performance attributes are the dimensions (length and width) of the quadcopter's arm, the motor power, the battery capacity, propeller diameters, and the payload. The lower and upper bounds of these parameters are shown in table \ref{tb:DVs}. 
We identify three unique talents based on the characteristics of the SWARM-CAE problem: search speed ($\mathbf{Y}_{\texttt{TL,1}}$), cruising speed ($\mathbf{Y}_{\texttt{TL,2}}$) and flight range ($\mathbf{Y}_{\texttt{TL,3}}$). For the search speed, we assume a linear correlation between the sensor and its weight; thus, the higher the payload, the higher the search speed. We executed the NSGA-2 multi-objective optimizer 6 times with an initial population of 120 and 40 generations each; the non-dominated samples from these runs were again filtered based on the non-dominated filtering process to get the final Pareto points.
For creating a Pareto model as explained in section \ref{subsec:talentpareto}, we considered search speed and the velocity to be independent variables and created a polynomial linear regression model to approximate the Fight Range.
The resulting model is shown in Fig~\ref{fig:paretomodel}. 
%For capturing the lower and upper bounds of the Pareto model, we used search speed $\hat{Y}_{\texttt{TL,1}}$ as an independent variable and predicted the 5th quantile for lower bounds ($L_{\hat{Y}_{\texttt{TL,2}}}$) and 95th quantile for upper bounds of cruising speed ($U_{\hat{Y}_{\texttt{TL,2}}}$).

\begin{table}
\centering
\caption{Talent Metrics and Design Variables of UAVs achieved in Co-design compared with Baseline Fixed Design}
\label{tb:DVs}
\includegraphics[trim={0 9.2cm 0 0},clip,width=0.35\textwidth]{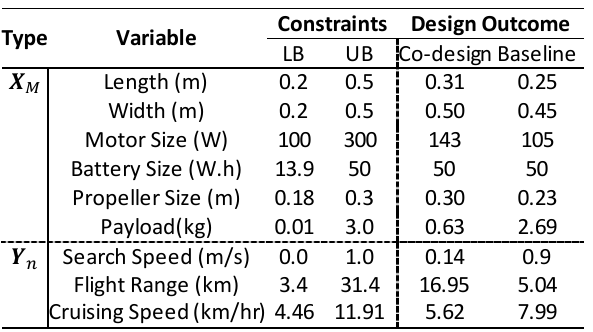}
      \end{table} 

\subsection{Behavior Learning subject to Talent Boundary}
\subsubsection{Policy Creation}
We used Stable-baselines3 \cite{stable-baselines3}, a standard open-source RL Library for creating a custom policy, distribution, and neural networks as discussed in section \ref{sec:GRL}. The policy outputs the search speed $\hat{Y}_{\texttt{TL,1}}$, cruising speed $\hat{Y}_{\texttt{TL,2}}$ and a behavioral action($a$). 
%The search speed and cruising speed are in a range of 0 to 1; since the bounds of search speed $\hat{Y}_{\texttt{TL,1}}$ are already between 0 and 1 \ref{tb:DVs}, no scaling is necessary. The quantile regression model then predicts the lower and upper bounds of the cruising speed $\hat{Y}_{\texttt{TL,2}}$ for the provided $\hat{Y}_{\texttt{TL,1}}$ and it is scaled based on equation \ref{eqn:scaleytl2}. 
\subsubsection{Training}
\begin{figure}[!ht]
    \centering
    \begin{subfigure}{0.49\linewidth}
        \includegraphics[width=\textwidth]{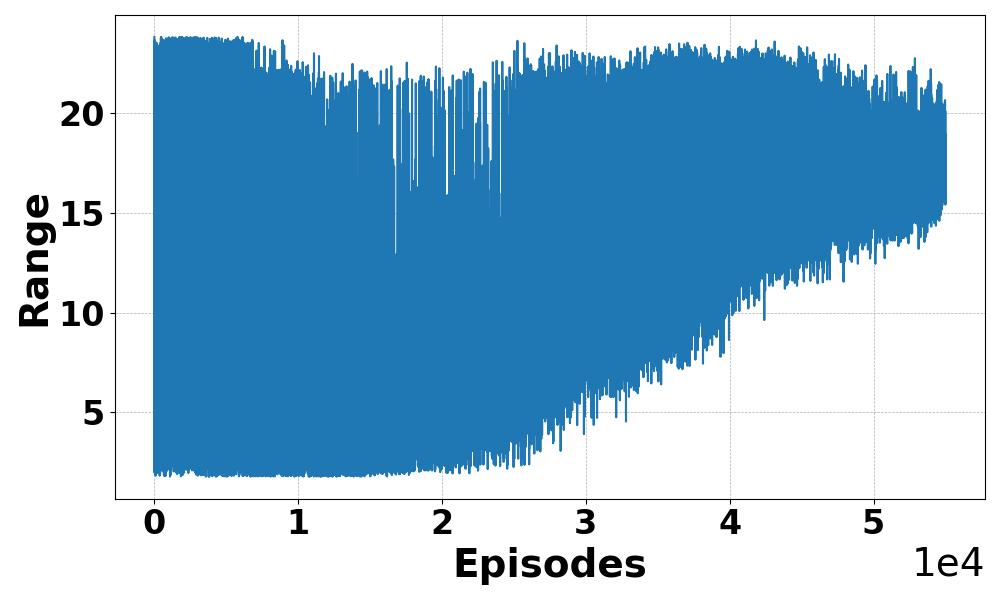}
        \caption{}
        \label{fig:range_convergence}
    \end{subfigure}
    \hfill
    \begin{subfigure}{0.49\linewidth}
        \includegraphics[width=\textwidth]{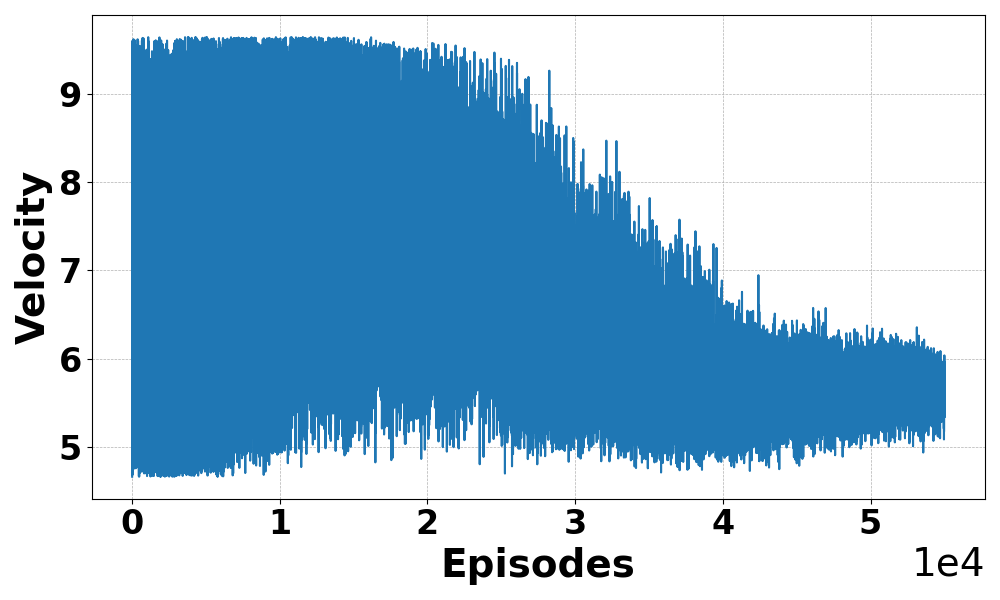}
        \caption{}
        \label{fig:velocity_convergence}
    \end{subfigure}
    \hfill
    \begin{subfigure}{0.49\linewidth}
        \includegraphics[width=\textwidth]{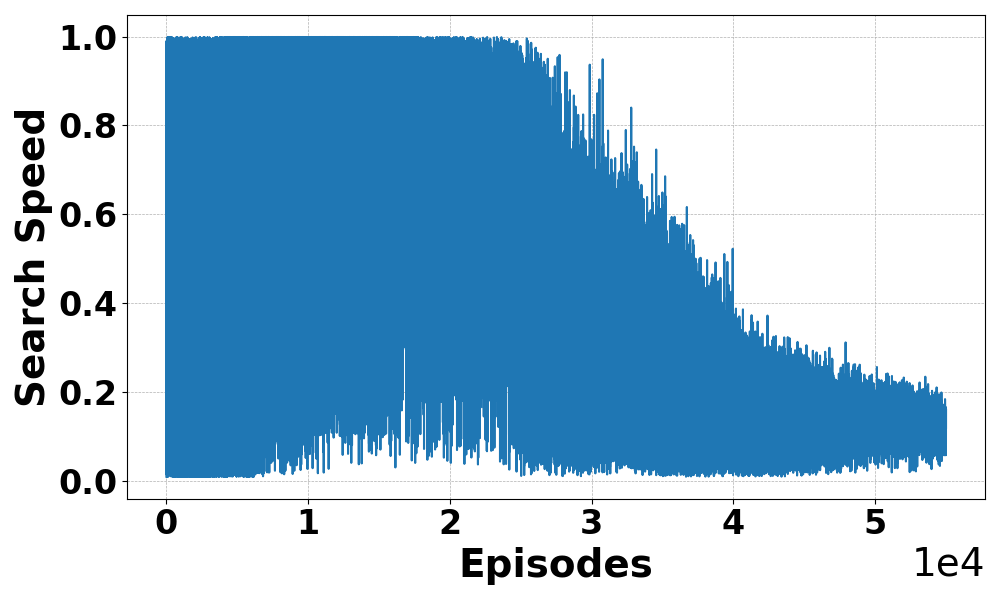}
        \caption{}
        \label{fig:searcspeed_convergence}
    \end{subfigure}
    \hfill
    \begin{subfigure}{0.49\linewidth}
        \includegraphics[width=\textwidth]{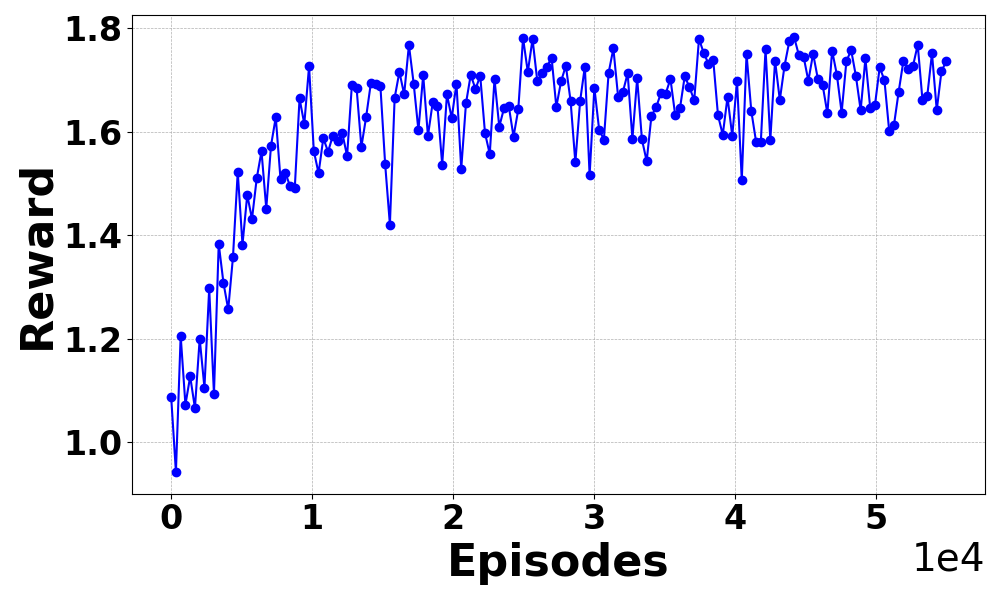}
        \caption{}
        \label{fig:reward_convergence}
    \end{subfigure}
    \caption{Training history (Talents and overall reward): a) Flight range b) Cruise speed, c) Search speed, d) Mission Rewards}
    \label{fig:overall_convergence}
\end{figure}

We trained the swarm tactics policy using the Talent-infused Actor-Critic method for 3 million timesteps simulated in the Buffalo Downtown region, keeping the platoons counts fixed at $N_{UAV}:4, N_{UGV}:4, N_{BLD}:10$, and $N_{ADV}:6$, with a consistent depot location. Even though every episode of training can have any combination of the above-mentioned parameters, we set it as constant during training since this enables us to stack the state space variables as tensors for faster training using GPUs. Each episode can be considered as a function evaluation with respect to the behavior and talents $f_L(\mathbf{Y}_{TL}, \boldsymbol\Phi) = R$, where R is the mission completion reward given by Eq.~\eqref{eq:rewards}. We introduced 30 unique scenarios, varying goals, robot numbers(they form as 4 platoons), target buildings, and adversaries, yet all scenarios began from the same depot. In each episode, a scenario was randomly selected from this pool, and the policy underwent training for 3 million timesteps with a learning rate of $1e-3$. We conducted parallel training (10 threads) on a 24-core server with 64 GB of memory. Figure \ref{fig:overall_convergence} displays convergence history for the three talent variables and rewards over 55,000 episodes. 

 Figure \ref{fig:overall_convergence} d) shows the rewards converge at around 22000 episodes. At the onset of training, the confidence level or the variance of the policy due to the Gaussian distribution in RL policy is high, and this allows for higher exploration in talent space. This is evident from Fig~\ref{fig:overall_convergence} a), b), and c). As the training progresses and the rewards get to a steady level, the confidence level increases, which reduces the variance in the talents. The final cumulative standard deviation in the trained policy after 55,000 episodes is 13\%. The policy enforces higher range, lower speed, and lower search speed. The training scenarios are created in such a way that the goal locations(victim's location) are at different distances from the depot location, and in order to be successful in all scenarios, the UAVs require a higher range. Due to its high range, it has to sacrifice its speed and/or payload. Since each UAV platoon consists of multiple robots and they collaboratively search different areas of the buildings, it is not necessary for an individual vehicle to have high-quality sensors leading to high payloads, and this explains the convergence of search speed to a low value. The total training time is approximately 160 hours, and hence, for each episode(single function evaluation of $f_L(\mathbf{Y}_{TL}, \boldsymbol\Phi)$), it takes ~10.47 seconds. 
The optimized talents $\mathbf{Y}_{\texttt{TL}}^*$ from the RL policy after the training are given in table \ref{tb:DVs}

\subsection{Morphology Finalization}
\begin{figure}[h!]
    \centering
    \includegraphics[width=0.6\linewidth]{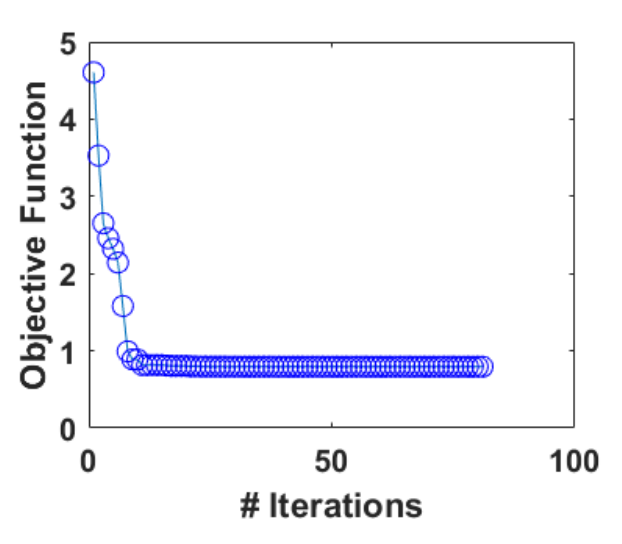}
    \caption{Morphology Finalization convergence history.}
    \label{fig:morphology_finalization}
\end{figure}
Using the optimized talents $\mathbf{Y}_{\texttt{TL}}^*$ we got from the training, We use Mixed-Discrete Particle Swarm Optimization (MDPSO)\cite{chowdhury2013mixed} to optimize for the best morphology. The objective here is to find suitable morphology that is as close as possible to the required talents. With an initial population of 150, the optimization ran for 80 iterations. The convergence history is shown in Fig~\ref{fig:morphology_finalization}.
The final morphology closely matched the learned talents, with an error under 0.9, demonstrating the effectiveness of the polynomial regression-based Pareto model (Table \ref{tb:DVs}). The optimization process took 110.8 seconds.

\begin{figure}[ht]
  \centering
  \begin{minipage}[b]{0.8\linewidth}
    \includegraphics[width=\linewidth]{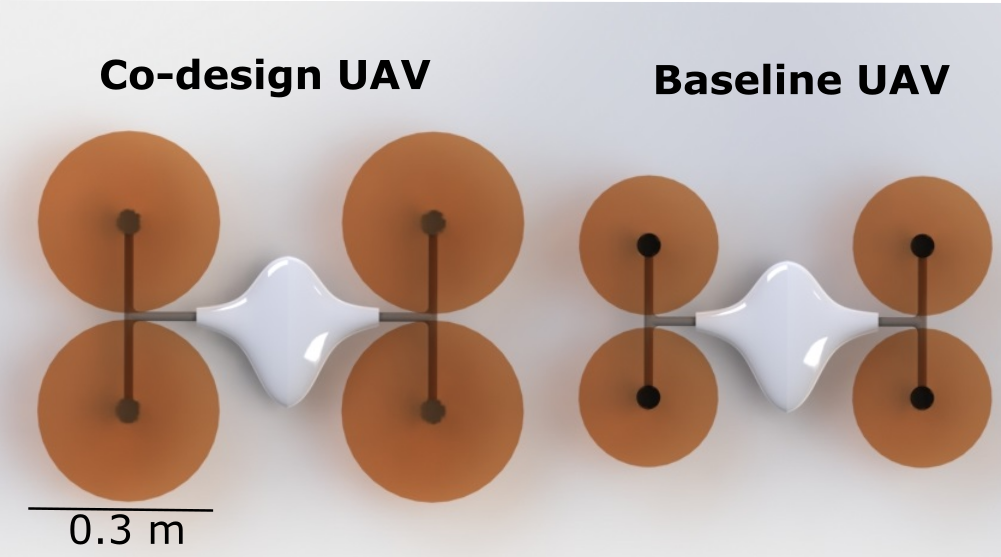}
  \end{minipage}
  \caption{CAD model showing the comparison of optimized UAV design (left) and the Baseline UAV (right); note that they also are significantly different in their motor size and payload capacity, which are not illustrated here.}
  \label{fig:cad_models}
\end{figure}

\subsection{Performance Analysis}

In this section, we compare the results of our co-designed policy with a fixed-design policy. To get a suitable talent for fixed-design policy, we calculated the average distance between the depot and target locations with a cautious path selection, which is 758 meters, and the distance between each target location is approximately 550 meters. There are a maximum of 8 target locations, and if a single UAV platoon decides to go to all 8 target locations, the maximum range it requires is around 5 KM. We randomly sampled a Pareto point from the Pareto solutions we got from section \ref{sec:sub_talents} with a 5 KM range. Note that obtaining values from Pareto points results from the sequential design process explained in section \ref{sec2}. We narrowed down a set of values based on environmental factors and selected the optimal Pareto value to ensure the most robust and effective comparison.
Since we are compromising on the range, we get higher search speed and cruising speed, allowing the UAV platoons to go to different locations and complete the search faster. Both the optimized talents and fixed talents are shown in Fig~\ref{fig:paretomodel} and in table \ref{tb:DVs}. We trained this RL policy for the same number of episodes as the co-design policy; note that here, the policy doesn't contain an additional talent network as we did for the co-design policy. The fixed design policy only outputs the action $a$ to be taken given the state $s$, whereas the talents remain fixed. The state space, reward, and all hyper-parameters are kept the same. 
\begin{figure}
    \centering
    \includegraphics[width=0.99\linewidth]{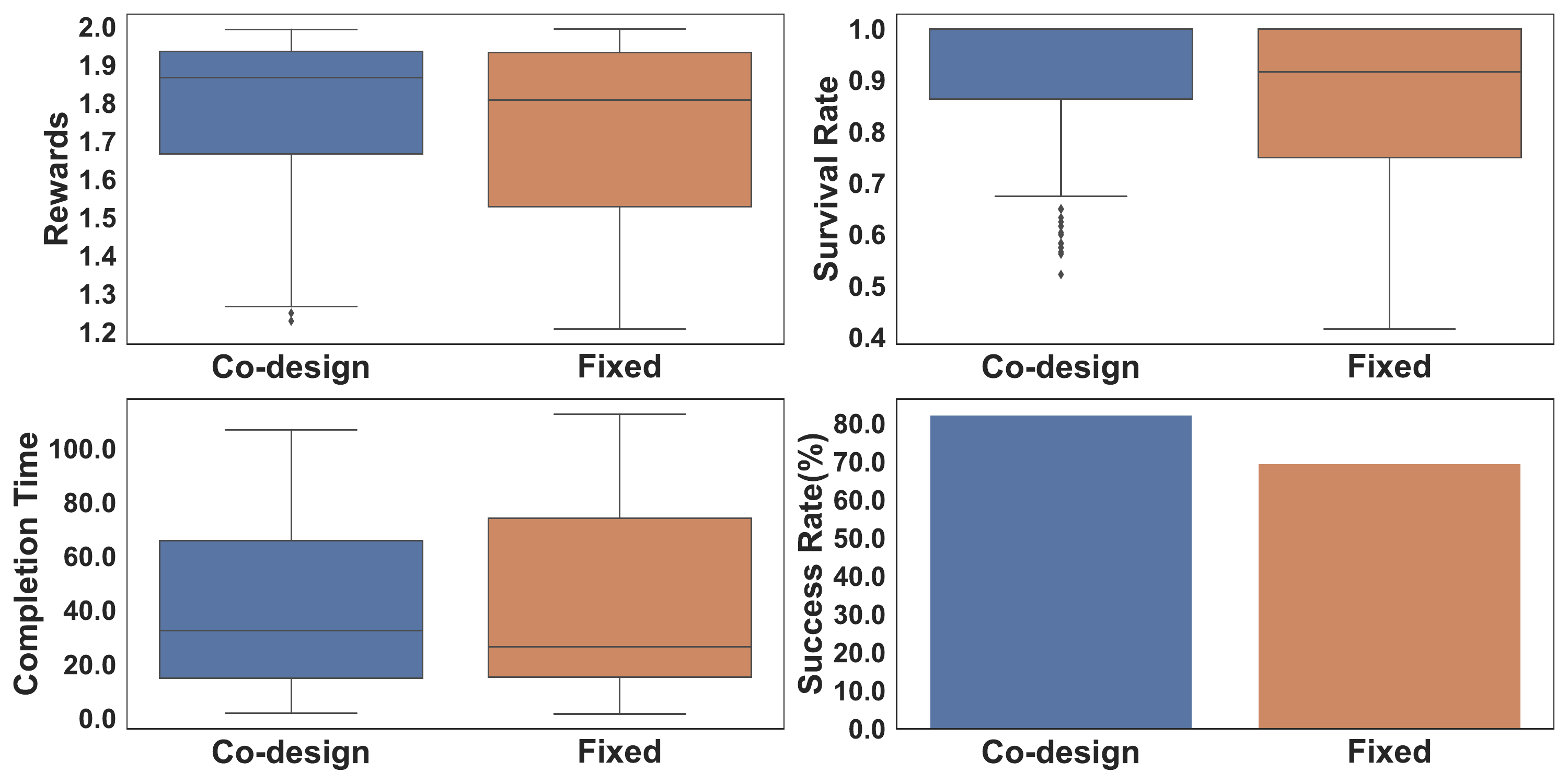}
    \caption{Performance of Co-design Policy and Fixed Baseline Trained Policy in training map. The metrics used for comparisons are a) Rewards, b) Survival rate (Number of remaining robots at end of mission), c) Completion Time of the rescue operation, and d) Success Rate of rescue operations (Bar plot showing total successful mission completion)}
    \label{fig:test_results}
\end{figure}

To evaluate the performance of the trained policies, we handcrafted an additional 40 distinct scenarios with high complexity for testing. These scenarios contain varying critical swarm parameters such as the number of robots, the locations of targets and goals, and the presence of adversaries. Each policy was tested for 250 episodes, and the outcomes are presented in Fig~\ref{fig:test_results}. As shown in the figure, the co-design policy achieves a success rate of about 82\%, whereas the fixed design policy achieves only 61\%. 
Our analysis focuses on 3 key metrics: total rewards, completion time, and survival rate. These metrics are only applicable for successful scenarios and hence don't reflect the performance. The total rewards, quantifying the cumulative reward achieved at the end of each episode, are computed as per Eq.~\eqref{eq:rewards}. Notably, the completion time and survival rate influence the reward metric. Hence, we also provide a comparison of these parameters. The co-design policy also has a higher reward and higher survival rate with less variance than that of a fixed-design policy. The co-design Policy has a higher completion time; this is primarily due to low cruising speed.   The CAD model comparing the Baseline UAV and Co-designed UAV is shown in Fig~\ref{fig:cad_models}.

\subsection{Computing Costs Analysis}
In this section, we compare the computational time taken by our proposed co-design framework to the nested co-optimization. 

Our talent-behavior co-optimization was trained in a workstation with Intel CPU-12900k (24 Threads), NVIDIA 3080ti, and 64 GB of RAM. The computation times for each step in our co-design framework are as follows: 6.7 minutes for 6 runs of NSGA-2 to obtain talent metrics, nearly negligible time (3.5 seconds) for creating a Pareto boundary, approximately 160 hours for talent-behavior actor-critic optimization using 20 parallel environments for experience collection, and 1.8 minutes for finalizing morphology. Overall, our co-design framework incurs a total computational cost of approximately 160.10 hours, with a significant portion of this time allocated to the learning process.

%Based on these computing costs, let's consider a fully-nested optimization for this problem with NSGA2 determining optimal morphologies, and for each morphology, we train the RL behavior for 20,000 episodes (this is based on the time it took to begin initial convergence). For the same NSGA2 settings that we used in our framework, i.e., population size of 120 and 40 generations, it will take approximately 2333 hours for a single run and 13,998 hours for 6 runs. This calculation is based on training all 120 behaviors in parallel, so the time for each generation is time per episode multiplied by 20,000, $10.47 \times 20000$, which is 58.33 hours. If each behavior learning uses 20 parallel environments, the workstation requires at least 240 threads. Note that this calculation is based on training all 120 populations in parallel, with each training using 20 parallel environments for experience collection. It requires a workstation with 240 threads, which is computationally inefficient.

For the same settings of NSGA-2, i.e., a population size of 120 and 40 generations, and considering each behavioral learning takes 20,000 episodes, a single run of NSGA-2 will take an estimated 2333 hours. This estimate is based on the assumption that all 120 behavioral learning happen in parallel, while each behavioral learning uses 20 parallel environments to collect experiences. 
With the nested co-optimization, there is a necessity to search the overall morphology space, whereas, in our co-design approach, the morphology-talent mapping and utilizing the pareto front for talent-behavior learning convert the morphology search space into Talent-Pareto search (search within the non-dominated solutions), which essentially makes our co-design framework extremely frugal in terms of computational time and computational hardware requirements compared to the nested co-optimization approach. 

% The 5 optimization trials for talent exploration took 32 minutes 6 seconds (1926 seconds), with a self time (cost of the algorithm) of 13 minutes 13 seconds; the Bayesian optimization trial for the CBM optimization took 3 hours 29 minutes 36 seconds (12,576 seconds), with a self time of 26 seconds; the optimization trial for morphology design costs 6 minutes 52 seconds; modeling the feasible region of the talent metrics costs less than 30 seconds to compute. The overall computing time of the entire case study is just over 4 hours (247 minutes).

% The average computing cost of morphology evaluation is negligible (around 0.12 second), the average cost of evaluating the CBM objective is around 83.0 seconds (with parallel computing). Based on the computing costs observed from the trials, the hypothetical computing cost of running a fully nested co-design optimization with the NSGA-II solver with identical settings is estimated as 14.6 days (350 hours). In comparison, our talent-based co-design optimization is extremely frugal on computing load.

%%%%% Conclusions %%%%%%%%%%%%%%%%%%%%%%%%%%%%%%%

\section{Conclusion} \label{sec:conclusion} 
In this paper, we introduce an efficient co-design framework to concurrently design the behavior and morphology by decomposing this optimization process into multiple search processes, the most critical among which is a talent-behavior co-learning process that is also constrained by a pre-computed talent Pareto. This process uses a novel Talent-infused Actor-Critic. To demonstrate the effectiveness of the proposed framework, we apply it to design the morphology and behavior of quadcopter type UAVs that are operating as a swarm along with a team of UGVs. Here, the behavior encompasses tactical decisions regarding tasks to allocate to different UAVs/UGVs in order to complete the mission in minimal time and with minimal loss of robots due to adversaries. These decisions are provided by the behavior policy model, trained by graph RL. Compared to a baseline sequential design (with morphology chosen from the talent Pareto and behavior learned separately), the co-design obtained outcome performs significantly better in terms of mission success rate. The overall co-design costs were also estimated to be 14 times smaller than what a nested co-optimization would have cost in terms of computing time. In its current form, the proposed approach hinges on the ability to identify talent metrics that are purely a function of morphology (i.e., independent of the control/behavior models). Hence, future work could explore autoencoders or related approaches to identify latent spaces to serve as the talent space instead and, therefore, allow the presented decomposition approach to work in a wider range of problems. 

\bibliographystyle{IEEEtran}
\bibliography{refs}

\end{document}